\documentclass[9pt]{ieeeconf}

\usepackage{bm} 
\usepackage{siunitx}
\robustify\bfseries
\usepackage{booktabs}
\usepackage[capitalize,nameinlink]{cleveref}
\usepackage{pifont}

\newcommand{\cmark}{\ding{51}}
\newcommand{\xmark}{\ding{55}}
\crefname{equation}{}{}
\Crefname{equation}{Equation}{Equations}

\title{Learning What to Degrade: A Recoverability-Aware Curriculum for Reliable Smart-Glasses Super-Resolution}

\addauthor{Yaopeng Wang}{1,3}
\addauthor{Qingliang Wang}{2,3}
\addauthor{Yilun Jin}{3}
\addauthor{Xiujin Liu}{4}
\addauthor{Han Yue}{3}
\addauthor{Chen Xia}{3}
\addauthor{Renjie Zhu}{3}
\addauthor{Chuang Sheng}{3}
\addaffiliation{1}{Southeast University}
\addaffiliation{2}{Fuzhou University}
\addaffiliation{3}{Ant Group}
\addaffiliation{4}{University of Michigan}

\begin{document}

\maketitle

\begin{abstract}

Degradations in smart-glasses images obscure small text and fine-grained structures needed by vision-language models (VLMs) for visual assistance. We introduce an high-resolution (HR)-referenced semantic evaluation protocol that compares VLM outputs on restored images with those on corresponding HR references. Our evaluation reveals two failure modes: generative SR can synthesize misleading details, whereas restoration-oriented SR can leave critical evidence blurred or ambiguous. These findings motivate improving faithful detail recovery in restoration-oriented SR, where a key training challenge is to provide sufficiently difficult inputs while retaining recoverable visual information. We propose ReDiC (Recoverability–Difficulty Curriculum), which combines a model-independent recoverability proxy with model-relative restoration difficulty to adapt training degradation severity. ReDiC increases severity when inputs are recoverable and restoration is easy, maintains challenging but recoverable conditions, and reduces severity when estimated recoverability is low. It modifies only training degradation generation, leaving the SR architecture and reconstruction objective unchanged. Across NAFNet, SwinIR, and MambaIR, ReDiC improves average PSNR by approximately 2.5 dB and HR-referenced semantic scores by about 3 points over the corresponding baselines on synthetic benchmarks, with further gains confirmed by LLM-based and human evaluations on 200 real smart-glasses captures.

\end{abstract}    
\section{Introduction}

\begin{figure*}[t]
    \centering
    \includegraphics[width=\linewidth]{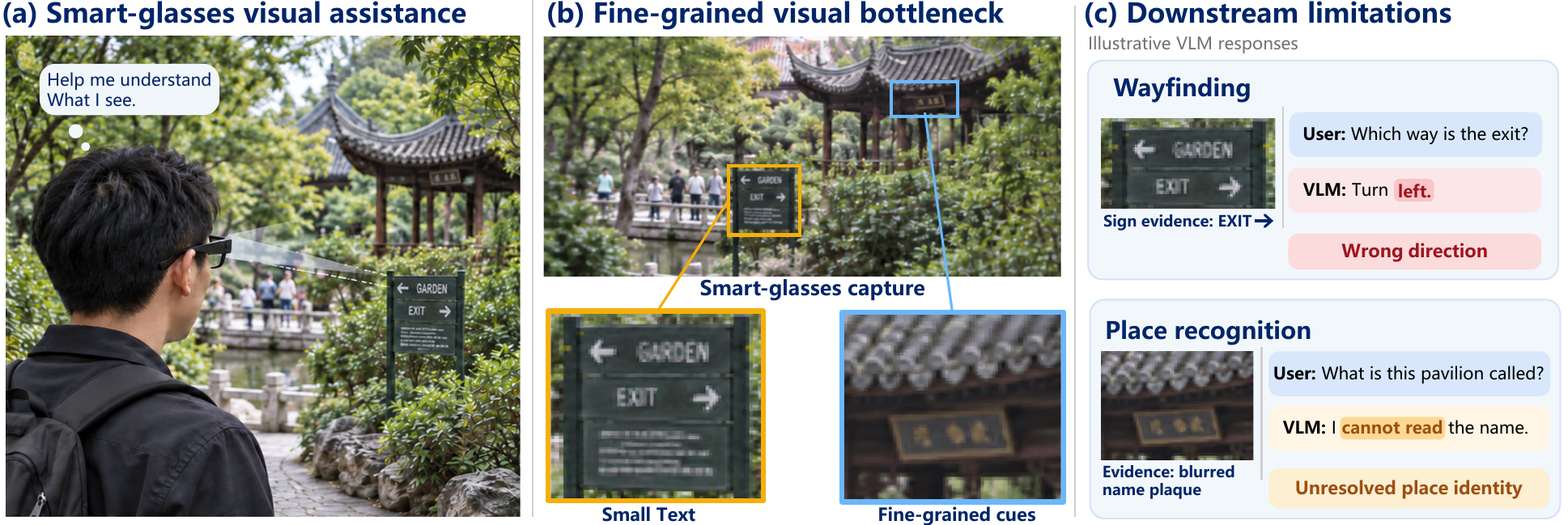}
    \caption{
    Visual evidence bottlenecks in smart-glasses assistance. 
    (a) A user wearing smart glasses for visual assistance in an everyday environment. 
    (b) A smart-glasses capture with enlarged regions highlighting small text and fine-grained visual cues obscured by limited image quality. 
    (c) Illustrative VLM responses show two potential downstream consequences: incorrect navigation guidance from a misread exit sign and failed place identification from an unreadable name plaque.
    }
    \label{fig:motivation}
\end{figure*}


Smart-glasses visual assistance relies on vision-language models (VLMs) to interpret visual in everyday scenes. 
However, captured images often suffer from low resolution, blur, and other degradations that obscure small text and fine-grained structures~\cite{engel2023project}. 
The loss of these cues can impair text recognition and scene understanding~\cite{wang2020scene,qiu2025benchmarking}, leading to errors such as misreading an exit sign or failing to identify a place, as illustrated in Figure~\ref{fig:motivation}. 
Super-resolution (SR) offers a means of recovering these details and improving the visual information available to downstream models.

Improved visual quality, however, does not necessarily lead to more reliable downstream understanding. Generative SR methods exploit learned priors to synthesize perceptually plausible details~\cite{wang2021real,lin2024diffbir}, which may be unsupported by the observed image and can mislead downstream VLMs~\cite{li2026evaluating}. 
To investigate this issue, we introduce a high-resolution (HR)-referenced semantic evaluation protocol that compares VLM outputs on restored images with those on corresponding HR references. Our evaluation reveals two failure modes: generative SR can introduce misleading details, whereas restoration-oriented SR can leave critical evidence blurred or ambiguous. 
Figure~\ref{fig:motivation} illustrates how altered or unresolved digits affect phone-number recognition. These findings motivate improving fine-grained detail recovery in restoration-oriented SR while preserving fidelity to observed content.

A key challenge in restoration-oriented SR is determining how to construct training inputs that remain sufficiently challenging without losing information needed for faithful recovery. 
Existing degradation pipelines, such as BSRGAN and Real-ESRGAN~\cite{zhang2021designing,wang2021real}, sample corruptions from predefined distributions but do not explicitly adapt severity to the model's learning progress. As the restoration ability of a model improves, previously challenging degradations may become easier to handle. Increasing degradation severity can maintain a useful restoration challenge, but may also remove visual information essential for accurate reconstruction. Reconstruction error provides a natural signal for estimating the restoration difficulty of the current inputs. However, high error may reflect either limited model capability or excessive information loss, requiring different adjustments to degradation severity. Challenging inputs that retain sufficient visual evidence provide useful learning opportunities, whereas inputs with low recoverability call for reducing degradation severity. 

We propose ReDiC (Recoverability–Difficulty Curriculum), which combines a model-independent recoverability proxy with model-relative restoration difficulty to adapt training degradation severity. 
The recoverability proxy estimates whether degraded inputs retain sufficient visual information for recovery, while relative difficulty assesses how challenging those inputs are for the current SR model. Patch-level estimates are aggregated and smoothed to guide three scheduling actions: increasing severity when inputs are recoverable and restoration is easy, maintaining the current severity under challenging but recoverable conditions, and reducing severity when estimated recoverability is low. ReDiC modifies only training-time degradation generation, leaving the SR architecture and reconstruction objective unchanged. Experiments across NAFNet, SwinIR, and MambaIR demonstrate improvements in restoration quality, text recognition, and HR-referenced semantic agreement on synthetic benchmarks, with automatic and human evaluations further confirming semantic improvements on 200 real smart-glasses captures.

Our contributions are summarized as follows:
\begin{itemize}
\item We introduce an HR-referenced semantic evaluation protocol that reveals two failure modes affecting downstream VLM understanding: misleading synthesized details and insufficient detail recovery.
\item We propose ReDiC, which jointly uses a model-independent recoverability proxy and model-relative restoration difficulty to increase, maintain, or reduce training degradation severity.
\item We demonstrate improvements across three SR backbones on synthetic benchmarks and 200 real-world smart-glasses captures through LLM-based and human evaluations.
\end{itemize}

\section{Related Work}
\label{sec:related_work}

\textbf{Real-World Super-Resolution and Degradation Modeling. }
Real-world SR relies on both restoration architectures and realistic degradation synthesis. BSRGAN~\cite{zhang2021designing} constructs complex degradations by randomly composing multiple operations, while Real-ESRGAN~\cite{wang2021real} adopts a high-order degradation process. RealDGen~\cite{peng2025towards} further learns realistic paired data by decoupling image content from degradation characteristics. These methods demonstrate the importance of degradation modeling, but the degradation distribution is typically predefined or learned independently of the current restoration state. In contrast, we treat degradation generation as a dynamic training process and adapt its severity according to both sample recoverability and model-relative restoration difficulty.

\textbf{Generative Priors and Semantic Fidelity.}
Generative SR leverages learned image priors to synthesize realistic details under severe degradation~\cite{ledig2017photo}. Representative methods include StableSR~\cite{wang2024exploiting}, DiffBIR~\cite{lin2024diffbir}, SeeSR~\cite{wu2024seesr}, and SUPIR~\cite{yu2024scaling}, which exploit pretrained diffusion priors, degradation-aware guidance, or large-scale generative models. Recent work also explores controllable fidelity--realism trade-offs~\cite{zhu2026one}. Despite their perceptual improvements, these methods may synthesize details unsupported by the LR observation. Meanwhile, DA-CLIP and related work explore vision-language representations for image restoration~\cite{luo2024controlling, li2026evaluating, wang2026cogsenet, wang2026cfsr}. Our focus is complementary: we investigate whether restored images provide reliable evidence for downstream VLMs.

\textbf{Curriculum Learning and Adaptive Difficulty.}
Curriculum learning improves optimization by organizing training examples according to difficulty~\cite{bengio2009curriculum, kumar2010self, graves2017automated}. Recent work on diffusion models explicitly estimates denoising difficulty across timesteps and constructs an easy-to-hard curriculum from the resulting training dynamics~\cite{kim2025denoising, xu2024profake}. However, in image restoration, stronger degradation does not necessarily correspond to more useful difficulty: once relevant information is destroyed, high reconstruction error may reflect intrinsic ambiguity rather than an informative hard sample~\cite{dong2015image}. We therefore distinguish model-relative restoration difficulty from degradation recoverability, and adapt degradation severity jointly according to whether a sample remains recoverable and whether the current model has mastered it.

\section{Semantic Reliability Gap in Super-Resolution}
\label{sec:semantic_reliability}
\begin{figure*}[t]
\centering
\includegraphics[width=\linewidth]{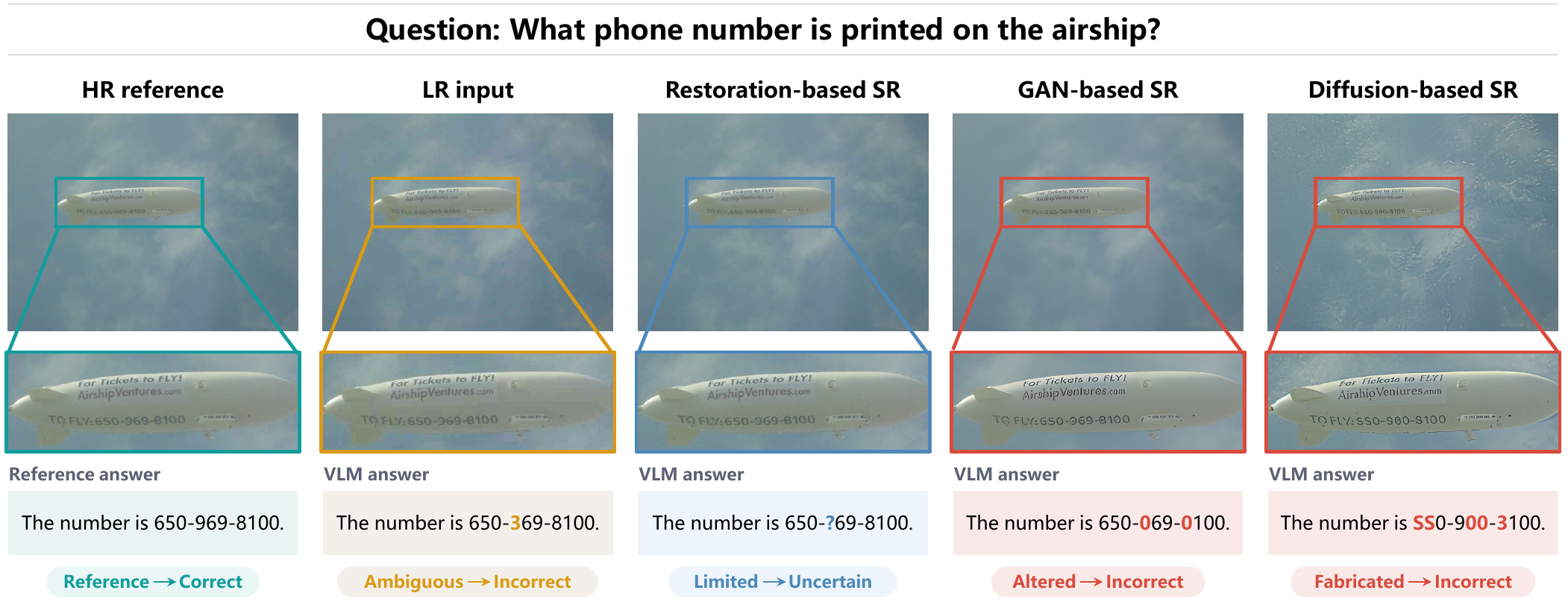}
\caption{
Visual appearance and downstream text understanding in a phone-number reading example. 
Columns show the HR reference, LR input, and outputs from restoration-, GAN-, and diffusion-based SR methods. 
Each column includes an enlarged phone-number region and the corresponding VLM response to the same question. 
The generative SR examples illustrate how sharper-looking text can contain altered digits and yield responses that disagree with the HR-derived answer.
}
\label{fig:sr_semantic_gap}
\end{figure*}

We examine whether SR improvements translate into more reliable downstream semantic understanding. Figure~\ref{fig:sr_semantic_gap} illustrates a phone-number reading example, comparing visual restoration with the corresponding VLM responses.

\subsection{HR-Referenced Semantic Evaluation}
\label{subsec:hr_semantic_eval}

For an HR reference $x^{\mathrm{HR}}$, its corresponding LR observation $x^{\mathrm{LR}}$, and an SR output $\hat{x}^{\mathrm{SR}}$, we apply the same fixed VLM $\Phi$ independently to all three images to generate image descriptions:
\begin{equation}
r=\Phi(x^{\mathrm{HR}}),\quad
d^{\mathrm{LR}}=\Phi(x^{\mathrm{LR}}),\quad
d^{\mathrm{SR}}=\Phi(\hat{x}^{\mathrm{SR}}).
\end{equation}
The HR-derived description $r$ serves as a semantic reference rather than manually annotated ground truth. Both LR and SR descriptions are independently evaluated against this common reference.

Following the LLM-as-a-Judge paradigm~\cite{zheng2023judging, liu2023g}, we employ four independent LLM judges: GLM-5, MiniMax-M2.1, Qwen3.5, and GPT-5~\cite{zeng2026glm, minimax2025m21, qwen2026qwen35, singh2025openai}. Each judge evaluates a candidate description against the HR-derived reference along four dimensions, assigning a score between 0 and 100 to each: semantic consistency, information coverage, factual fidelity, and semantic relevance. These dimensions assess agreement with key facts, retention of reference information, contradictions or unsupported claims, and alignment with task-relevant details, respectively.

For judge $m$, the aggregate score is computed as
\begin{equation}
\begin{aligned}
S^{(m)}(d,r) ={}&
0.4S_{\mathrm{con}}^{(m)}(d,r)
+0.25S_{\mathrm{cov}}^{(m)}(d,r)\\
&+0.25S_{\mathrm{fid}}^{(m)}(d,r)
+0.1S_{\mathrm{rel}}^{(m)}(d,r).
\end{aligned}
\end{equation}
All judges receive identical evaluation instructions and structured output formats, and use low-temperature decoding. We collect their dimension-wise scores and compute the weighted aggregate externally. The final score is the average across the three judges:
\begin{equation}
S(d,r)=\frac{1}{3}\sum_{m=1}^{3}S^{(m)}(d,r).
\end{equation}

We further define the semantic gain of an SR method over the LR input as
\begin{equation}
\Delta S =
S(d^{\mathrm{SR}},r)-S(d^{\mathrm{LR}},r).
\label{eq:semantic_gain}
\end{equation}
Positive $\Delta S$ indicates improved agreement with the HR-derived reference, while negative values indicate reduced agreement. Since the reference is VLM-generated rather than manually verified, these scores measure semantic agreement under the fixed evaluation protocol rather than absolute factual correctness.


\subsection{Perceptual Enhancement Does Not Guarantee Semantic Reliability}
Figure~\ref{fig:sr_semantic_gap} provides a qualitative phone-number reading example. The same question is presented to the VLM for the HR reference, LR input, and each restored image. The responses illustrate how altered digits in generative SR outputs can lead to incorrect interpretations, whereas unresolved digits in restoration-oriented SR outputs can remain difficult to recognize.

To quantitatively examine this discrepancy, we use paired synthetic degradation data, where HR images are degraded and subsequently restored by the evaluated SR methods. PSNR and LPIPS are computed against the original HR references, while semantic evaluation follows Sec.~\ref{subsec:hr_semantic_eval}, using HR-derived descriptions as common references. As shown in Table~\ref{tab:motivation}, Real-ESRGAN achieves the lowest LPIPS, while DiffBIR obtains the highest PSNR, yet their semantic scores decrease by 5.71 and 2.27 points relative to the LR input, respectively. In contrast, restoration-oriented methods achieve semantic gains of 0.24--0.84 points. These results demonstrate that improved reconstruction or perceptual quality does not necessarily translate into better semantic agreement. This motivates improving faithful detail recovery in restoration-oriented SR, where the key challenge is to construct sufficiently difficult training inputs without sacrificing recoverable visual information. ReDiC addresses this challenge by jointly considering model-independent recoverability and model-relative restoration difficulty to adapt training degradation severity.

\begin{table}
\vspace{-12pt}
\centering

\caption{
Reconstruction quality and HR-referenced semantic agreement on synthetically degraded images. PSNR and LPIPS are computed against the corresponding HR references, while Judge Score measures semantic agreement with HR-derived references.
$\Delta$ Score denotes the change relative to the LR input.
Best results in each metric column are bold.
}
\label{tab:motivation}

\footnotesize
\setlength{\tabcolsep}{2.2pt}
\renewcommand{\arraystretch}{1.10}

\resizebox{0.98\linewidth}{!}{%
\begin{tabular}{@{}lccccc@{}}
\toprule
\textbf{Method}
& \textbf{Type}
& \textbf{PSNR}$\uparrow$
& \textbf{LPIPS}$\downarrow$
& \textbf{Judge}$\uparrow$
& $\Delta$\textbf{ Score}$\uparrow$ \\
\midrule

LR
& --
& --
& --
& 65.75
& -- \\

\midrule

NAFNet
& Restoration
& 22.93
& 0.4684
& 66.46
& +0.71 \\

SwinIR
& Restoration
& 22.94
& 0.4793
& 65.99
& +0.24 \\

MambaIR
& Restoration
& 22.96
& 0.4761
& \textbf{66.59}
& \textbf{+0.84} \\

\midrule

Real-ESRGAN
& GAN
& 22.97
& \textbf{0.2945}
& 60.04
& -5.71 \\

DiffBIR
& Diffusion
& \textbf{23.29}
& 0.3273
& 63.48
& -2.27 \\

\bottomrule
\end{tabular}%
}

\vspace{-6pt}
\end{table}

\section{Method}
\label{sec:method}
\vspace{-4pt}
\subsection{Overview}
\label{sec:overview}

\begin{figure*}[t]
    \centering
    \includegraphics[width=\textwidth]{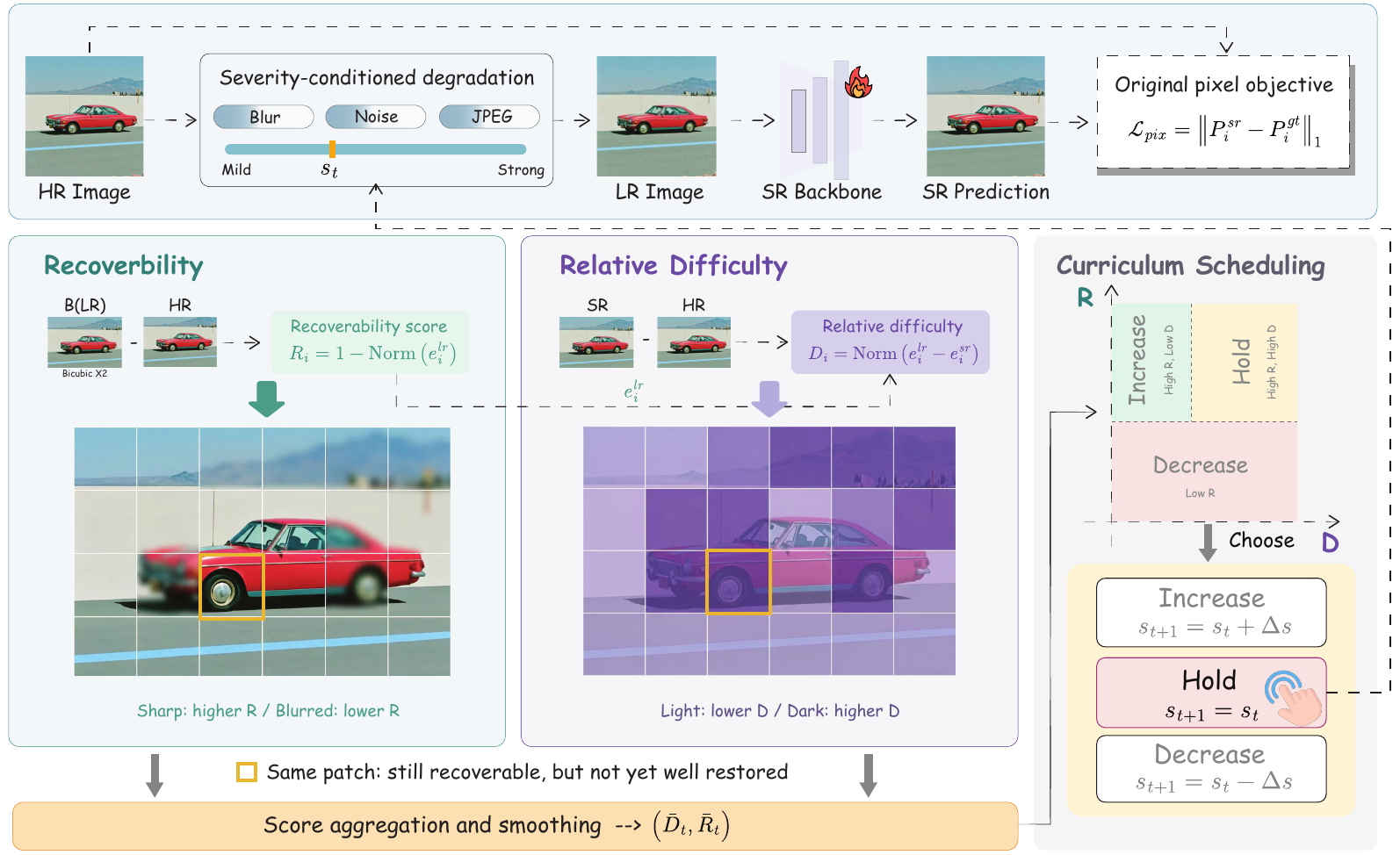}
    \caption{
        Overview of ReDiC. A severity-conditioned degradation
        process generates LR training samples with adaptive
        difficulty. ReDiC estimates recoverability and relative
        difficulty from the generated samples, and a curriculum
        scheduler adjusts degradation severity according to the
        resulting learning state. The SR architecture and
        reconstruction objective remain unchanged.
    }
    \label{fig:framework}
\end{figure*}

Figure~\ref{fig:framework} summarizes ReDiC: severity-conditioned degradation generates LR inputs, and recoverability–difficulty estimates guide subsequent severity updates. The curriculum changes only degradation generation, leaving the SR architecture and reconstruction objective unchanged.

\subsection{Severity-Conditioned Degradation}
\label{sec:degradation}

Given an HR image $I^{\mathrm{gt}}$, we generate its LR observation
at curriculum window $t$ as
\begin{equation}
I^{\mathrm{lr}} = \mathcal{G}(I^{\mathrm{gt}}; \phi_t),
\qquad
\phi_{k,t} \sim p_k(\phi_k \mid s_t),
\end{equation}
where $\mathcal{G}$ combines blur, noise, compression, and resampling,
$\phi_t = \{\phi_{k,t}\}_{k=1}^{K}$ denotes the sampled degradation
parameters, and $s_t \in [0,1]$ is the curriculum-controlled severity.
The parameter sampling ranges are continuously interpolated between
predefined mild and strong configurations.
Increasing $s_t$ therefore shifts the degradation distribution toward
stronger corruption while preserving stochastic diversity among samples.
The SR scale remains fixed at $\times 2$, so the curriculum adjusts
degradation severity without changing the reconstruction scale.

\subsection{Recoverability--Difficulty Learning-State Estimation}
\label{sec:state}


\textbf{Degradation recoverability.}
For an LR patch $P_i^{\mathrm{lr}}$ with HR target
$P_i^{\mathrm{gt}}$, we obtain a fixed reconstruction by bicubic
upsampling, $P_i^{\mathrm{b}}=\mathcal{B}(P_i^{\mathrm{lr}})$,
and compute its error
$e_i^{\mathrm{lr}}
=|\Omega_i|^{-1}\|P_i^{\mathrm{b}}-P_i^{\mathrm{gt}}\|_1$,
where $\Omega_i$ denotes the patch support.
We define the recoverability score as
\begin{equation}
R_i
=
1-
\operatorname{clip}
\left(
\frac{e_i^{lr}-e^{low}}
     {e^{high}-e^{low}+\epsilon},
0,1
\right),
\label{eq:recoverability}
\end{equation}
where $e^{\mathrm{low}}$ and $e^{\mathrm{high}}$ are fixed
empirical calibration bounds.
A larger $R_i$ indicates lower reconstruction error under fixed
interpolation, providing a model-independent proxy for
degradation recoverability.

\vspace{0.3em}
\noindent\textbf{Relative restoration difficulty.}
Given the SR prediction
$P_i^{\mathrm{sr}}=F_{\theta}(P_i^{\mathrm{lr}})$,
we compute its reconstruction error
$e_i^{\mathrm{sr}}
=|\Omega_i|^{-1}\|P_i^{\mathrm{sr}}-P_i^{\mathrm{gt}}\|_1$
and normalize it by the interpolation baseline:
\begin{equation}
D_i
=
\operatorname{clip}
\left(
\frac{e_i^{sr}}
     {\max(e_i^{lr},e^{low})},
0,1
\right).
\label{eq:difficulty}
\end{equation}

Normalizing by the interpolation error expresses the model’s remaining error relative to baseline distortion. High $R_i$ and high $D_i$ indicate relatively low baseline distortion but limited error reduction by the model. Low $R_i$ indicates large baseline distortion and motivates reducing degradation severity.

\vspace{0.3em}
\noindent\textbf{Robust state aggregation.}
To obtain a stable learning-state estimate, we aggregate patch-level scores at both image and curriculum-window levels. Let $\mathcal{P}_j$ denote the sampled patches from image $j$. We first compute image-level scores using quantile aggregation:
\begin{equation}
R_j = Q_{q_R}(\{R_i\}_{i\in\mathcal{P}_j}),
\qquad
D_j = Q_{q_D}(\{D_i\}_{i\in\mathcal{P}_j}),
\end{equation}
where $Q_q(\cdot)$ denotes the $q$-th quantile.

For a curriculum window $t$ containing $N_t$ training
images, we average their image-level scores:
\begin{equation}
\tilde{R}_t =
\frac{1}{N_t}\sum_{j=1}^{N_t}R_j,
\qquad
\tilde{D}_t =
\frac{1}{N_t}\sum_{j=1}^{N_t}D_j.
\end{equation}
We then apply exponential moving averages to reduce
short-term fluctuations:
\begin{equation}
\begin{aligned}
\bar{R}_t &= \beta\bar{R}_{t-1}
+(1-\beta)\tilde{R}_t,\\
\bar{D}_t &= \beta\bar{D}_{t-1}
+(1-\beta)\tilde{D}_t,
\end{aligned}
\end{equation}
where $\beta\in[0,1)$ is the smoothing coefficient.
For the first curriculum window, we initialize
$\bar{R}_0=\tilde{R}_0$ and $\bar{D}_0=\tilde{D}_0$.
The resulting state $(\bar{R}_t,\bar{D}_t)$ is used
by the curriculum scheduler.

\subsection{Recoverability-Aware Curriculum Scheduling}
\label{sec:scheduler}

At curriculum window $t$, ReDiC updates degradation severity
using the smoothed state $(\bar{R}_t,\bar{D}_t)$:
\begin{equation}
s_{t+1} =
\begin{cases}
\operatorname{clip}(s_t-\Delta s,0,1),
& \bar{R}_t < \tau_R,\\
\operatorname{clip}(s_t+\Delta s,0,1),
& \bar{R}_t \geq \tau_R
  \;\land\; \bar{D}_t < \tau_D,\\
s_t,
& \text{otherwise},
\end{cases}
\end{equation}
Here, $\Delta s>0$ denotes the fixed severity adjustment
step, while $\tau_R$ and $\tau_D$ are the recoverability
and difficulty thresholds, respectively. The scheduler
is initialized with severity $s_0$ and updates $s_t$
once per curriculum window containing $N_t$ training
images. Low recoverability triggers a severity reduction,
whereas severity increases only when recoverability is
sufficient and restoration difficulty is low. Otherwise,
the current severity is maintained.
Training retains the original $\ell_1$ reconstruction objective:
\begin{equation}
\mathcal{L}_{\mathrm{pix}}
=
\frac{1}{B}\sum_{i=1}^{B}
\left\|
F_{\theta}(P_i^{\mathrm{lr}})-P_i^{\mathrm{gt}}
\right\|_1.
\end{equation}


\section{Experiments}
\label{sec:experiments}

\begin{table*}[t]
    \centering
    \caption{
        Quantitative comparison of restoration and downstream
        performance under synthetic degradations.
         Avg. denotes the mean across three severity levels, and $\Delta$ Score denotes the Judge Score gain over LR Input.
    }
    \label{tab:main_results}

    \small
    \setlength{\tabcolsep}{5pt}
    \renewcommand{\arraystretch}{1.15}

    \begin{tabular*}{\textwidth}{
        @{\extracolsep{\fill}}lcccccccc@{}
    }
        \toprule
        \textbf{Method}
        & \shortstack{\textbf{Mild}\\PSNR$\uparrow$}
        & \shortstack{\textbf{Medium}\\PSNR$\uparrow$}
        & \shortstack{\textbf{Hard}\\PSNR$\uparrow$}
        & \shortstack{\textbf{Avg.}\\PSNR$\uparrow$}
        & \shortstack{\textbf{Avg.}\\LPIPS$\downarrow$}
        & \shortstack{\textbf{Avg.}\\OCR-NED$\downarrow$}
        & \shortstack{\textbf{Avg.}\\Judge$\uparrow$}
        & \shortstack{\textbf{Avg.}\\$\Delta$ Score$\uparrow$}
        \\
        \midrule

        LR Input
        & -- & -- & -- & --
        & -- & 0.5490 & 71.04 & --
        \\

        \midrule

        NAFNet
        & 30.92 & 29.36 & 25.44 & 28.57
        & 0.2825 & 0.5503 & 73.54 & 2.50
        \\

        NAFNet + ReDiC
        & 33.97 & 31.41 & 27.42 & 30.93
        & 0.2500 & 0.4860 & 76.47 & 5.43
        \\

        \midrule

        SwinIR
        & 31.02 & 29.48 & 25.46 & 28.65
        & 0.2905 & 0.5369 & 73.83 & 2.79
        \\

        SwinIR + ReDiC
        & 34.17 & 31.63 & 27.34 & 31.05
        & 0.2361 & 0.4564
        & 76.87 & 5.83
        \\

        \midrule

        MambaIR
        & 31.19 & 29.54 & 25.46 & 28.73
        & 0.2918 & 0.5390 & 73.96 & 2.92
        \\

        MambaIR + ReDiC
        & 34.39
        & 31.91
        & 27.83
        & 31.38
        & 0.2298
        & 0.4372
        & 76.85 & 5.81
        \\

        \bottomrule
    \end{tabular*}
\end{table*}

\subsection{Experimental Setup}
\label{sec:exp_setup}

\textbf{Datasets.}
We use 800 DIV2K~\cite{agustsson2017ntire,timofte2017ntire} and 4,000 LSVT~\cite{sun2019icdar} images for training, with 100 DIV2K and 500 LSVT validation images for synthetic evaluation. HR images are degraded at three severity levels: Mild ($s\in[0.2,0.4]$), Medium ($s\in[0.4,0.6]$), and Hard ($s\in[0.6,0.8]$), with the original images serving as pixel-aligned references. We also evaluate on 200 real-world smart-glasses captures with separately photographed HR references from matched viewpoints.

\noindent\textbf{Evaluation metrics.}
For synthetic data, we report PSNR and LPIPS~\cite{zhang2018unreasonable} against HR references, and OCR-NED on text-containing LSVT images using PaddleOCR~\cite{cui2026pp} outputs from HR images as reference texts. Downstream semantic understanding is evaluated using the HR-referenced LLM-as-a-Judge protocol in Sec.~\ref{sec:semantic_reliability}, reporting Judge Score and its gain over LR inputs ($\Delta$ Score). For real-world captures, we report LLM Judge Scores and human ratings. Their rank agreement is measured using Spearman's rank correlation coefficient over individual image-output pairs.

%
\noindent\textbf{Implementation details.}
We fine-tune NAFNet~\cite{chen2022simple}, SwinIR~\cite{liang2021swinir}, and MambaIR~\cite{guo2024mambair} from their pretrained checkpoints using online degradations and an $\ell_1$ reconstruction loss. Aligned $64\times64$ LR and $128\times128$ HR patches are extracted with paired rotations for augmentation, using a batch size of 16 per GPU. All models are trained for 30,000 iterations with Adam ($\beta_1=0.9$, $\beta_2=0.99$), an initial learning rate of $2\times10^{-4}$, and MultiStepLR milestones at 7,500, 15,000, and 22,500 iterations with a decay factor of 0.5. Training is conducted on NVIDIA A100 GPUs with FP16 precision. The degradation pipeline includes blur, noise, JPEG compression, and resampling. Baseline and ReDiC models share identical training data, initialization, and optimization settings, differing only in degradation scheduling.
For ReDiC, we set $s_0=0.20$, $\Delta s=0.08$, $\tau_R=0.36$, and $\tau_D=0.80$. The recoverability calibration bounds ($e^{\mathrm{low}}=0.018$, $e^{\mathrm{high}}=0.115$) correspond to the 5th and 95th percentiles of bicubic reconstruction errors on training patches. Curriculum statistics are updated every 200 images per GPU using 16 patches per image, with $q_R=0.25$, $q_D=0.75$, and EMA coefficient $\beta=0.80$. Statistics are aggregated across 16 GPUs, with the first valid window initializing the EMA.


\begin{figure*}[t]
    \centering
    \includegraphics[width=\linewidth]{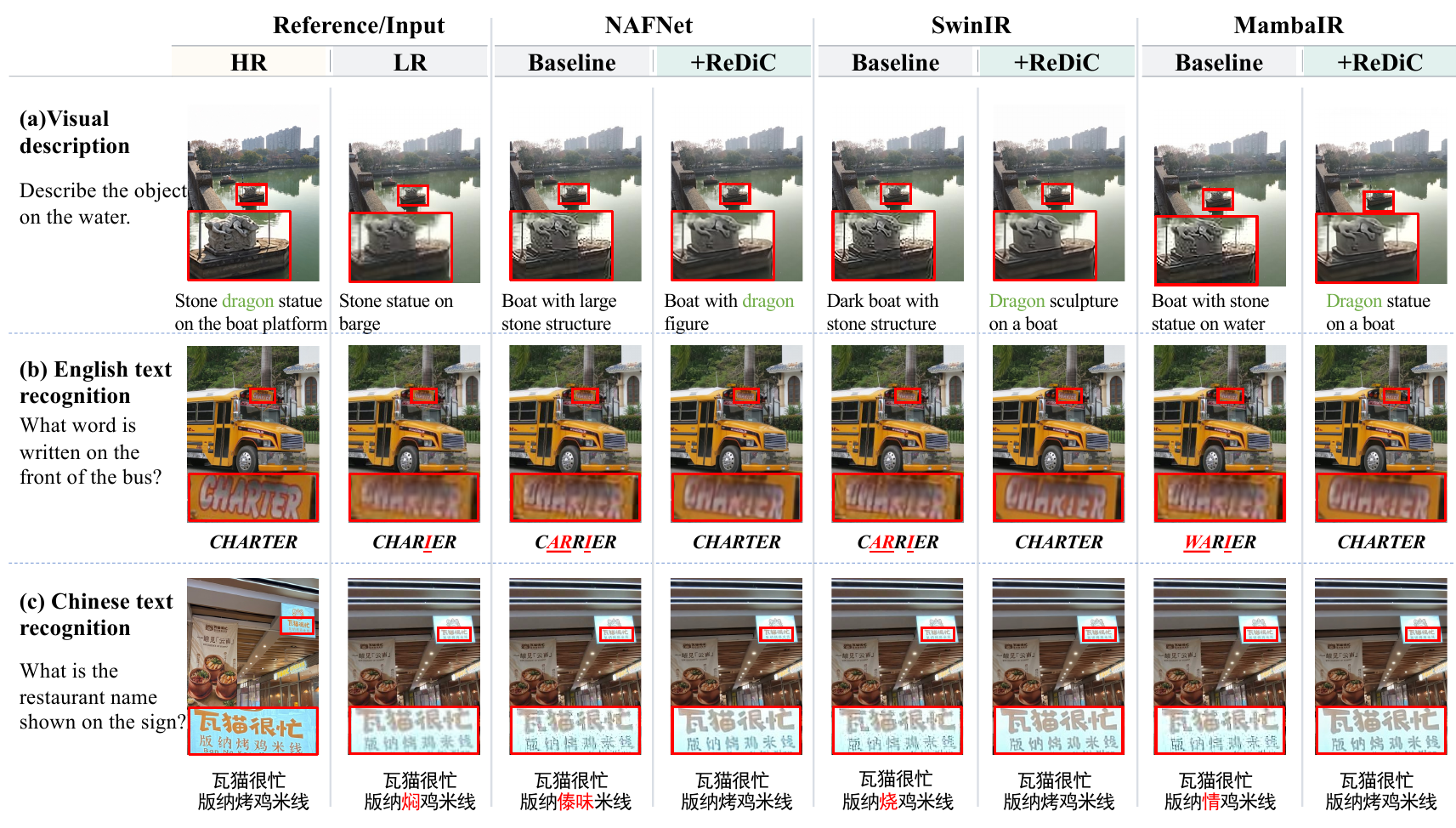}
\caption{
Qualitative comparison of SR outputs and corresponding LLM responses across NAFNet, SwinIR, and MambaIR, with and without ReDiC. Examples show (a) visual description, (b) English text recognition, and (c) Chinese text recognition. HR-derived responses serve as semantic references. Red boxes indicate enlarged regions.
}
\label{fig:semantic_gap}
\end{figure*}

\subsection{Performance under Synthetic Degradations}
\label{sec:main_results}

\textbf{Restoration quality.}
Table~\ref{tab:main_results} shows that ReDiC improves PSNR across all three backbones and degradation levels, with average gains of 2.36–2.65 dB over the corresponding baselines. Gains remain 1.88–2.37 dB under Hard degradation, showing that the benefit extends to more severely degraded inputs. Average LPIPS also decreases for every backbone, indicating improved perceptual similarity alongside higher reconstruction fidelity. 


\noindent\textbf{Downstream understanding.}
ReDiC improves both text recognition and HR-referenced semantic agreement across all three backbones. 
Average OCR-NED decreases by 11.7--18.9\% relative to the corresponding baselines, with MambaIR improving from 0.5390 to 0.4372. 
Average Judge Score increases by 2.89--3.04 points over the corresponding baselines. All ReDiC variants also outperform the LR input on both metrics, with Judge Score gains of 5.43--5.83 points. 
\subsection{Real-world Evaluation}
\label{sec:real_world}

\begin{table}
\vspace{-12pt}
\centering

\caption{
LLM-based and human evaluation on 200 real-world smart-glasses captures using HR-derived descriptions as references. Cons., Cov., Fid., and Rel. denote consistency, coverage, fidelity, and relevance. LLM Judge is their weighted aggregate, and Human Score is the mean rating on a 1--5 scale.
}
\label{tab:real_results}

\footnotesize
\setlength{\tabcolsep}{2.5pt}
\renewcommand{\arraystretch}{1.10}

\resizebox{0.98\linewidth}{!}{%
\begin{tabular}{@{}lcccccc@{}}
\toprule
\textbf{Method}
& \textbf{Cons.}$\uparrow$
& \textbf{Cov.}$\uparrow$
& \textbf{Fid.}$\uparrow$
& \textbf{Rel.}$\uparrow$
& \textbf{LLM Judge}$\uparrow$ 
& \textbf{Human Score}$\uparrow$ 
\\
\midrule

Raw Glasses
& 60.8 & 59.5 & 50.9 & 66.6 & 58.6 & --\\

Bicubic
& 60.6 & 59.7 & 49.9 & 66.6 & 58.3 & --\\

\midrule

NAFNet
& 60.3 & 60.9 & 50.2 & 66.5 & 58.5 & 3.46\\

NAFNet + ReDiC
& 64.1 & 61.3 & 52.2 & 69.0 & 60.9 & 3.63\\

\midrule

SwinIR
& 60.7 & 59.4 & 51.1 & 67.1 & 58.6 & 3.61\\

SwinIR + ReDiC
& 63.1 & 63.2 & 53.2 & 68.9 & 61.2 & 3.87\\

\midrule

MambaIR
& 59.6 & 60.5 & 50.1 & 67.0 & 58.2 & 3.56\\

MambaIR + ReDiC
& 63.0 & 63.5 & 52.3 & 69.3 & 61.1 & 3.77\\

\bottomrule
\vspace{-25pt}
\end{tabular}%
}

\end{table}


We evaluate ReDiC on 200 real-world smart-glasses captures covering text-rich scenes, distant objects, and fine-grained visual structures. 
Following Sec.~\ref{sec:semantic_reliability}, responses generated from the corresponding HR references serve as semantic references for evaluating the original and restored images.

\noindent\textbf{LLM-based Semantic Evaluation.}
We follow the same LLM inference and LLM judging configurations described in Sec.~\ref{sec:semantic_reliability}. 
We compare Raw Glasses, Bicubic, and three SR backbones with and without ReDiC. Table~\ref{tab:real_results} reports the results across four semantic dimensions and their weighted aggregate. ReDiC consistently improves the overall Judge Score across all three backbones, yielding gains of 2.4, 2.6, and 2.9 points for NAFNet, SwinIR, and MambaIR, respectively. Improvements are also observed across the individual semantic dimensions, indicating that ReDiC enhances the semantic information recovered from real-world smart-glasses images.
\noindent\textbf{Human Evaluation.}
We conduct a human evaluation with 20 participants on the 200 real-world captures, with each participant evaluating 50 images and each image receiving five independent ratings. For each image, participants are presented with the HR reference image and three pairs of candidate LLM outputs, each corresponding to the baseline and ReDiC-restored images of one SR backbone. They rate each output using the same four evaluation dimensions as the LLM-based evaluation, with a 5-point Likert scale. Method and backbone identities are concealed, and candidate presentation orders are randomized. As shown in Table~\ref{tab:real_results}, ReDiC improves human evaluation scores across all three backbones. Furthermore, LLM-based scores show strong rank correlation with mean human ratings, achieving a Spearman rank correlation coefficient of 0.81 across 1,200 image-output pairs.

\begin{figure*}[!t]
\centering
\includegraphics[width=\linewidth]{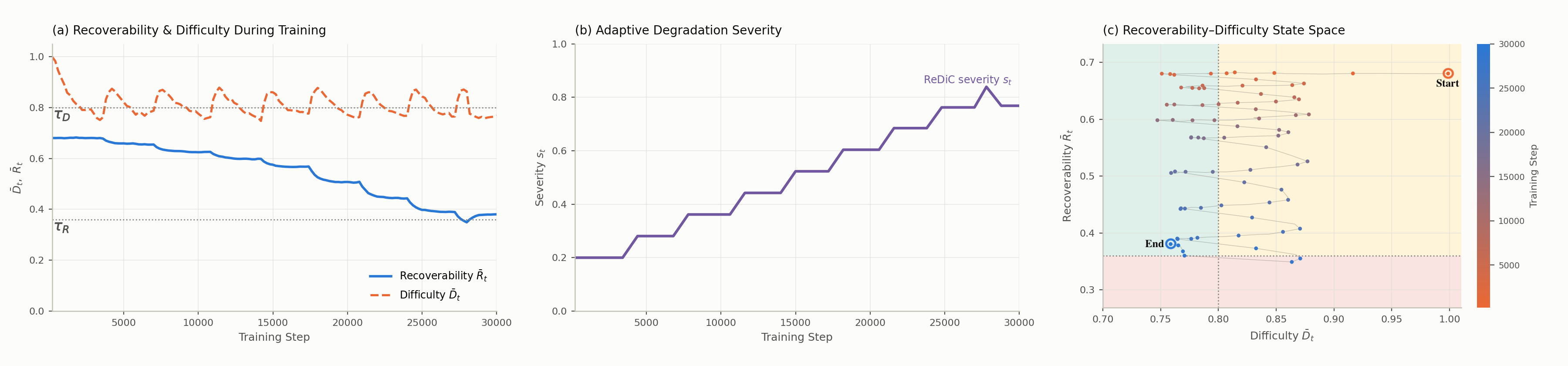}
\caption{Training dynamics of ReDiC. (a) Recoverability $\bar{R}_t$ and relative difficulty $\bar{D}_t$. (b) Degradation severity $s_t$. (c) Trajectory in the $(\bar{D}_t,\bar{R}_t)$ space. Dashed lines indicate thresholds; colored regions denote severity increase, maintenance, and reduction.
}
\label{fig:curriculum_dynamics}
\end{figure*}

\subsection{Qualitative Analysis}
\label{subsec:qualitative_analysis}
Figure~\ref{fig:semantic_gap} compares restored images and their corresponding LLM responses, using HR-derived responses as references. In Figure~\ref{fig:semantic_gap}(a), all three ReDiC variants identify the dragon figure omitted by LR and baseline responses. In (b), they correctly recognize \texttt{CHARTER}, which is misread by the LR input and baseline methods. In the real-world example in (c), all methods recognize the main restaurant name, but only ReDiC variants recover the smaller accompanying text. These examples illustrate improvements in object identification and fine-grained text recognition, complementing the quantitative results.

\subsection{Ablation Study}
\label{sec:ablation}
We compare eight degradation strategies on MambaIR. The baseline independently samples each degradation operator; Random uses a shared severity sampled from $\mathcal{U}(0,1)$, while Fixed sets $s=0.5$. Easy-to-hard progressively increases severity from 0 to 1. R-only follows this schedule but holds severity once $\bar{R}_t<\tau_R$. D-only increases severity when $\bar{D}_t<\tau_D$ and $\bar{R}_t\geq\tau_R$, and otherwise holds it. Shuffled-ReDiC permutes ReDiC's severity sequence, while ReDiC jointly uses both signals to increase, maintain, or reduce severity.

Table~\ref{tab:ablation_rd} shows that the MambaIR baseline outperforms Random because independent sampling allows diverse degradation combinations, whereas shared severity couples operator strengths. Fixed also limits training diversity, while Easy-to-hard improves all metrics but cannot account for evolving model capability and input recoverability. Both R-only and D-only outperform the baseline: D-only achieves higher PSNR, whereas R-only performs better on OCR-NED and Judge, demonstrating the complementary roles of the two signals. ReDiC achieves the best results across all metrics, while the lower performance of Shuffled-ReDiC despite identical severity values highlights the importance of update order.

\begin{table}[t]
    \centering
    \caption{
        Ablation of degradation scheduling and the contributions of recoverability ($R$) and relative difficulty ($D$) on MambaIR. Results are averaged across three severity levels. Sch. denotes the scheduler.
    }
    \label{tab:ablation_rd}

    \footnotesize
    \setlength{\tabcolsep}{2pt}
    \renewcommand{\arraystretch}{1.15}

    \begin{tabular*}{\columnwidth}{
        @{\extracolsep{\fill}}lcccccc@{}
    }
        \toprule
        \textbf{Variant}
        & $\mathbf{R}$
        & $\mathbf{D}$
        & \textbf{Sch.}
        & \textbf{PSNR}$\uparrow$
        & \textbf{OCR-NED}$\downarrow$
        & \textbf{Judge}$\uparrow$
        \\
        \midrule

        MambaIR
        & \xmark & \xmark & \xmark
        & 28.73 & 0.5390 & 73.96
        \\

        \midrule

        Random
        & \xmark & \xmark & \xmark
        & 26.42 & 0.5980 & 70.80
        \\

        Fixed
        & \xmark & \xmark & \xmark
        & 27.85 & 0.5620 & 72.60
        \\
         Easy-to-hard
        & \xmark & \xmark & \xmark
        & 28.95 & 0.5280 & 74.18
        \\

        Shuffled-ReDiC
        & \cmark & \cmark & \xmark
        & 28.46 & 0.5480 & 73.55
        \\

        $D$-only
        & \xmark & \cmark & \cmark
        & 30.24 & 0.4932 & 74.85
        \\

        $R$-only
        & \cmark & \xmark & \cmark
        & 29.64 & 0.4733 & 75.42
        \\

        \midrule

        \textbf{ReDiC (Ours)}
        & \cmark & \cmark & \cmark
        & \textbf{31.38}
        & \textbf{0.4372}
        & \textbf{76.85}
        \\

        \bottomrule
    \end{tabular*}
\end{table}

\subsection{Curriculum Dynamics}
\label{sec:curriculum_dynamics}
Figure~\ref{fig:curriculum_dynamics} illustrates the evolution of recoverability, relative restoration difficulty, and degradation severity during training. The difficulty $\bar{D}_t$ increases following severity updates and decreases as the model adapts, while recoverability $\bar{R}_t$ generally declines under stronger degradations. Late in training, $\bar{R}_t$ falls below $\tau_R=0.36$ at $s_t\approx0.84$, triggering a reduction to approximately $0.76$, after which recoverability rises above the threshold. These dynamics demonstrate ReDiC's ability to adapt severity through progression, maintenance, and retreat rather than monotonic scheduling.


\section{Conclusion}
\label{sec:conclusion}

We investigated semantic reliability in smart-glasses SR using an HR-referenced evaluation protocol, revealing that improved visual quality does not necessarily lead to more reliable downstream understanding: generative SR can introduce misleading details, whereas restoration-oriented SR can leave critical details blurred or ambiguous. To improve detail recovery in restoration-oriented SR, we proposed ReDiC, which adapts training degradation severity by combining a model-independent recoverability proxy with model-relative restoration difficulty. ReDiC modifies only training-time degradation generation, leaving the SR architecture and reconstruction objective unchanged. Experiments across NAFNet, SwinIR, and MambaIR demonstrate average PSNR gains of 2.36--2.65~dB over the corresponding baselines under synthetic degradations, alongside improved text recognition and HR-referenced semantic agreement. On 200 real-world smart-glasses captures, LLM-based and human evaluations further confirm semantic improvements across all three backbones.


\bibliographystyle{IEEEtran}
\bibliography{refs}








\end{document}